\documentclass[11pt,letterpaper]{article}
\usepackage{emnlp2017}
\usepackage{times}
\usepackage{latexsym}
\usepackage{graphicx}

\usepackage{tabularx}
\usepackage{subcaption}
\usepackage{booktabs}
\usepackage{multirow}
\usepackage{graphicx}
\usepackage{amsmath}
\usepackage{comment}
\usepackage{pgfplots,wrapfig}
\usepackage{color}
\usepackage{xcolor,colortbl}
\usepackage{enumitem}
\usepackage{ amssymb }

\newcommand{\ra}[1]{\renewcommand{\arraystretch}{#1}}
\newcolumntype{P}[1]{>{\centering\arraybackslash}p{#1}}
\pgfplotsset{compat=1.12}

\emnlpfinalcopy

\def\emnlppaperid{***}

\title{Linguistic Reflexes of Well-Being and Happiness in Echo}

 \author{Jiaqi Wu, Marilyn Walker, Pranav Anand \and Steve Whittaker \\
         \\
 University of California Santa Cruz \\
  {\tt \{jwu64,mawalker,panand,swhittak\}@ucsc.edu}}

\begin{document}

\maketitle

\begin{abstract}
Different theories posit different sources for feelings of well-being
and happiness.  Appraisal theory grounds our emotional responses in
our goals and desires and their fulfillment, or lack of fulfillment.
Self-Determination theory posits that the basis for well-being rests
on our assessments of our competence, autonomy and social
connection. And surveys that measure happiness empirically note that
people require their basic needs to be met for food and shelter, but
beyond that tend to be happiest when socializing, eating or having
sex. We analyze a corpus of private micro-blogs from a well-being
application called {\sc echo}, where users label each written post
about daily events with a happiness score between 1 and 9.  Our goal
is to ground the linguistic descriptions of events that users
experience in theories of well-being and happiness, and then examine
the extent to which different theoretical accounts can explain the
variance in the happiness scores.  We show that
recurrent event types, such as {\sc obligation} and {\sc
  incompetence}, which affect people's feelings of well-being are not
captured in current lexical or semantic resources.

\end{abstract}

\section{Introduction}

There has recently been huge interest in well-being, with a recent
review arguing that psychological well-being plays a causal role in
promoting job success, physical health, and long-term relationships
\cite{Lyubomirskyetal05,Kahneman99}. In this paper we analyze a
corpus of private micro-blogs from a well-being application 
called {\sc echo},  with the aim to 
detect, understand, and further advance systems that can 
improve both short and longer-term issues with well-being.

{\sc Echo} initiates user-written reactions to daily events, called
{\sc recordings}, as well as subsequent {\sc reflections} on those
events at points in the future \cite{Isaacsetal13}.\footnote{The ECHO
  corpus is not publicly available because of the ethical agreement
  with ECHO users.  To protect users' privacy, the uploaded images 
 are not stored for analysis.} Each reaction is labelled {\it
    at the time of recording or reflection} by the {\it user}, the
  first-person experiencer, with a {\it happiness} rating from 1 and
  9. Note that all users' posts and ratings are private,
  distinguishing this corpus from public sources like LiveJournal,
  where the content of posts might be influenced by considerations of
  self-presentation. Figure~\ref{fig:echo-example} shows a {\sc
    recording} and {\sc reflection} from two users, after binning the
  happiness ratings into positive and negative.

\begin{figure}[t]
\small
\begin{tabularx}{\columnwidth}{X}
\toprule
{\sc recording} (\emph{Negative}): I have to clean the kitchen since it's my chore this week, but I really don't want to do it! \\
{\sc reflection} (\emph{Positive}): I'm glad I did it!! The kitchen was clean and I watched the kardashians while doing it! \\
\midrule 
{\sc recording} (\emph{Positive}): I am having a lovely lunch with my two friends. We are eating at Pacific Thai. Tom yuumm!! \\
{\sc reflection} (\emph{Negative}): I miss hanging out with friends, I've been so busy lately. \\
\bottomrule
\end{tabularx}
\caption{{\sc recording} and {\sc reflection} of Echo}
\label{fig:echo-example}
\end{figure}

Our goal is to ground the linguistic descriptions of events that users
experience, such as those in Figure~\ref{fig:echo-example}, in
theories of well-being and happiness.  Without such a grounding, it is
difficult for the {\sc echo} system to make recommendations to users
to improve their well-being, or to explain the relationships between
different event types and well-being, or to develop a policy
that can do a good job of selecting events
for targeted reflection \cite{konrad2015finding,Isaacsetal13}. That is,
for \textsc{echo}'s purposes, we need techniques that not only reliably
categorize a user's scalar happiness level, but are explanatory with respect
to the sources of that happiness level.

There are two principal challenges to this goal. First,
 different theories posit different sources for feelings of
well-being and happiness.  Second, the relevant
computational resources for sentiment or mood are primarily lexically based, while
many of the events can only be characterized well via their
compositional semantics \cite{ReschkeAnand11}. 

Other research also shares our motivation of understanding the
relationship between what people say and their levels of happiness and
related moods.   \newcite{mishne05} used a corpus of
340,000 posts from Livejournal that were self-annotated with the 40
most common moods. Lexical features alone improved classification
accuracy by 6 to 15\% over a balanced baseline.
These results were
then improved considerably
\cite{keshtkar2009using}. \newcite{mihalcealiu06} experimented with the
subset of  \emph{happy/sad} posts, and
used conditional probability to explore the ``happiness factor'' of
various terms, and the relationship of
these terms to well-being categories such as human-centeredness and
socialness. \newcite{Schwartzetal16} extract 5,100 public status updates
on Facebook and have Turkers annotate them using Seligman's
dimensions for well-being: Positive Emotions, Engagement, Relationships, Meaning, and Accomplish \cite{Seligmanetal06,Forgeardetal11}. They then predict each dimension
with lexical and LDA topic features.

A related line of work builds lexico-semantic resources for sentiment
analysis with a focus on how the participants of an event are affected
by it. \newcite{GoyalRiloff13} bootstrap a set of patient-polarity
verbs from narratives and \newcite{DingRiloff16} extract event-triples
from blogs that reliably indicate positive or negative affect on one
of the event participants. \newcite{Reedetal17} take a similar
approach.  Deng et al. \shortcite{Dengetal13} annotate how
participants of an event are affected, and Deng \& Wiebe
\shortcite{DengWiebe14} show that this assists inference about the
author's sentiment towards entities or events. \newcite{Balahuretal12}
use the narratives produced by the ISEAR questionnaire
\cite{SchererWallbott86} for first-person examples of particular
emotions (``I felt angry when X and then Y happened'') and extract
sequences of subject-verb-object triples, which they then annotate for
seven basic emotions.  Choi \& Wiebe \shortcite{ChoiWiebe14} use
WordNet to try to learn similar patterns, and Rupenhofer \& Brandes
\shortcite{RuppenhoferBrandes15} annotate synsets in GermaNet based on
an event decomposition framework.
\newcite{russo-caselli-strapparava:2015:SemEval} proposed a shared
task for recognition of a set of pleasant and unpleasant events from a
clinical framework for well-being \cite{macphillamy1982pleasant}.
Work on AFINN, SentiWordNet and the Connotation Lexicon also aim to
refine existing sentiment resources to capture more subtle notions of
sentiment
\cite{Fengetal13,Kangetal14,baccianella2010sentiwordnet,nielsen2011new}.

Here we report an exploratory study where we
synthesize theoretical constructs associated with well-being and
happiness from different sources.  We then develop several methods for
characterizing events in terms of these theories. We examine the
extent to which different theoretical accounts can explain the
variance in the happiness scores in {\sc echo}.  We show that each
theory explains a part of the variance, but that our event
characterizations need to be more fine-grained.  We show that several
recurrent event types which affect people's feelings of well-being,
such as {\sc obligation} and {\sc incompetence}, are not captured in
current lexical or semantic resources.

\begin{table*}[ht!]
%\begin{scriptsize}
\begin{small}
\centering
\begin{tabular}{c|c|l|c|p{2.75in}}
\toprule
Row \# & \bf Source  & \bf Subtype & \bf Affect  & \bf Example  \\
\midrule
1 & \bf Goals & Achieved & POS & I applied to an scholarship got a large chunk of my reading done and got started cramming for next test .\\
2 &       & Thwarted & NEG & Wasn't able to get back in time for my class section .\\
\midrule
3 & \bf Eudaimonics & Autonomy  & POS & Good day at work had the right support and students were listening and behaving which was awesome. \\
4 &             & Lack-Autonomy  & NEG & Long list of things to do before going out tonight. \\
5 &             & Competence & POS & After working hard and spending so many countless hours, I finally finished my project for my psych class ! \\
6 &            & Incompetence & NEG & My midterm was really long and I didn't finish. \\
7 &             & Connection & POS & Having a nice time with my parents watching the Opening Winter Olympic Ceremony.\\
8 &             & Lack or Neg-Connection & NEG & My friend needs a bone marrow biopsy and chemo. \\
\midrule
9 & \bf Hedonics  & Savouring  & POS &  I love home cooking! Especially if it's Italian. \\
  &                & Savouring  & NEG &  The bus was rather packed and had a few people bump into me from where I was sitting. \\
\bottomrule
\end{tabular}
%\end{scriptsize}
\end{small}
\caption{\label{theory-examp} Examples of Theoretical Categories and Instantiations in {\sc Echo}}
\end{table*}

\section{Background and Motivation}
\label{theory-sec}

\begin{figure}[ht!]
\centering
\includegraphics[width=2.2in]{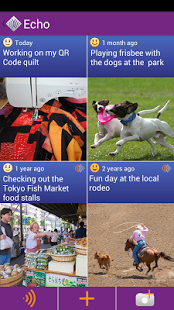}
\caption{Screenshot of the Echo Interface \label{echo-interface-fig}}
\end{figure}

{\sc Echo} is designed to encourage users to react to daily events as
well as to periodically reflect on past events
\cite{Isaacsetal13}. Figure~\ref{echo-interface-fig} depicts the user
interface, showing a {\sc recording} from today, as well as prompts to
reflect on events from the past.  {\sc echo} has been deployed with
134 users, in three different experiments on well-being
\cite{Konradetal16,Konradetal-TOCHI16}.  The total corpus consists of
10354 posts, where 7573 are {\sc recordings} and 2781 are {\sc
  reflections}. While the corpus could be considered relatively small,
these posts provide a window onto users' private thoughts as opposed
to what users are willing to make public on social media. In addtion,
the annotations for happiness are provided by the user, the
first-person experiencer, and not by a third party.

Our aim is to explain users' emotional reactions to different
categories of events mentioned in {\sc echo} posts, linking the
user reactions directly to theories of well-being as exemplified in
Table~\ref{theory-examp}. 

Influential accounts such as Appraisal Theory
\cite{scherer2001appraisal,SchererWallbott86,Ortonyetal90} argue that
success or failure in personal goals directly mediates affect.  Rows 1
and 2 in Table~\ref{theory-examp}.  Such mediation arises because
emotions have an important adaptive signaling function that serves to
motivate future behaviors in relation to those goals.  Row 1 provides
a description from {\sc echo} of successfully achieving
goals. Appraisal theory posits that goal achievement promotes positive
affect, which then serves to reinforce the relevant behavior.
Row 2 provides an example of failing to achieve an important
personal goal, which is posited to promote
negative affect, motivating people to modify current behaviors to
change that negative outcome.

There are significant critiques of the adaptive goal-based
account espoused in Appraisal theory.  Appraisal theory focuses on
short-term personal goals, but Eudaimonic psychologists instead focus
on what determines long-term happiness.  Eudaimonic theorists
suggest that certain fundamental psychological needs have to be
satisfied for people to experience sustained positive long-term
emotions. Self-determination theory argues that there are 3 basic
psychological needs: {\sc autonomy}, {\sc competence} and {\sc
  connection} \cite{DeciRyan10,RyanDeci00a,Bandura77}.  We add these to our
inventory in Table~\ref{theory-examp} in Rows 3 to 8. According to
self-determination theory, satisfaction of these basic needs results
in positive emotions.  Row 3 describes a good day at work. Row 5
describes feeling competent because hard work led to an achievement,
and Row 7 describes feeling connected with family. On the other hand,
if these basic needs are not satisfied, then negative emotions will
regularly arise. For example, obligations to do things one does not
feel like doing (Row 4), or a job that does not engage personal
decision making or involvement (lack of autonomy) can make one feel
unhappy. Similarly, people may feel unhappy due to an experience where
the demands of the situation outstrip one's basic abilities, such as
doing poorly on a test (lack of competence), as in Row 6.  In
addition, bad things happening to friends (Row 8) as well as
separation from family or friends often reduces happiness (lack of
connection).

In addition, there is strong evidence from {\sc savouring} theory
\cite{Joseetal12,bryant2011understanding} arguing that people often
experience highly positive or negative emotions arising from
situations that aren't directly goal-related, and that relate more
directly to basic drives \cite{maslow1943theory,Elson-Thesis12}. For
example, experiences such as eating, experiencing nature, sex and
physical exercise tend to engender positive emotions, whereas pain,
discomfort and inactivity have the opposite effects, and these are
documented in results from happiness surveys
\cite{Kahnemanetal04,Seligmanetal06}.  Thus while experiences such as
eating may serve the survival goal of preventing starvation, avoiding
starvation is unlikely to be a direct personal goal every time we eat,
suggesting that such experiences are not explained by Appraisal
theory. Similar arguments have been made by Lewinsohn and colleagues
who have shown that encouraging people to engage in certain simple
activities (shopping, mowing the lawn, driving, personal hygiene) have
quite predictable effects on mood without engaging significant
personal goals
\cite{macphillamy1982pleasant,Lewinsohnetal85,Lewinsohnetal78}.

\section{Empirical Approach}
\label{method-overview-sec}

\begin{table}[h!t]
\footnotesize
\centering
\begin{tabularx}{2.55in}{p{0.7in}| c c c  }
\toprule
\bf  Dataset & \bf Pos & \bf Neg & \bf Total  \\
\midrule
Train &  4743 &  3180 & 7923  \\
Test &  810 & 515 & 1325 \\ 
\bottomrule
\end{tabularx}
\caption{\label{tab:data-split} Number of Sentences for Train and Test}
\end{table} 

We start with the 10354 posts from the {\sc echo} corpus 
and map happiness scores between [1, 4] to negative,
and scores between [6, 9] to positive. For posts labelled 5 by the
experiencer, we categorize it as negative if its {\sc reflection} score
decreases to lower than 5, and positive if its {\sc reflection} score
increases. We label the rest of the 5s as neutral, and leave them
aside. We then have 5997 positive posts and 3573 negative posts. We
randomly sample 2868 posts as training data, and 478 as test
data. We keep the rest of the 6224 posts untouched for future work. Then we split the posts into sentences. Table~\ref{tab:data-split} shows the splits for each class.

\begin{table}[h!t]
\footnotesize
\centering
\begin{tabularx}{2.55in}{p{0.55in}| c| c c  c}
\toprule
\bf Features & \bf Metric & \bf Pos & \bf Neg & \bf All  \\
\midrule
UniGram &  Prec &  0.75  & 0.66 &  0.72  \\
        &  Rec & 0.81 & 0.59  & 0.72 \\ 
        &  F1 &  0.78 & 0.62 & 0.72 \\
\midrule
LIWC &  Prec & 0.72 & 0.72 & 0.72  \\
     &  Rec & 0.89 & 0.45 & 0.72  \\
     &  F1 & 0.80 & 0.55 &  0.70 \\
\bottomrule
\end{tabularx}
\caption{\label{tab:baseline-SVM-results} Weighted Metrics for SVM on Test}
\end{table}  

 \begin{table}[h!t]
% \footnotesize
 \small
 \centering
 \begin{tabularx}{2.55in}{p{0.75in}|c}
 \toprule
 \bf Unigram & \bf LIWC \\
 \midrule
 fun& affect,posemo,leisure \\
 good  & affect,posemo,drives,reward \\
 we &  we,social,drives,affiliation \\
 lunch &  bio,ingest \\
 glad &  affect,posemo  \\
 want & cogproc,discrep \\
 why & interrog,cogproc,cause \\
 need & cogproc,discrep \\
 no & negate \\
 not & negate,cogproc,differ \\
 \bottomrule
 \end{tabularx}
 \caption{\label{tab:unigram} The most informative UniGram features weighted by Information Gain} 
 \end{table}

\begin{table}[h!]
 \small
%\footnotesize
\centering
 \begin{tabularx}{2.55in}{p{0.75in}|c}
 \toprule
 \bf LIWC & \bf Words \\
 \midrule
 negemo & stress*, sad, sick, hate  \\
 posemo & fun, well, great, love \\
 negate & dont, didnt, no, cant, havent\\
 anger& hate, frustrat*, annoying \\
 i & i, my, me, im, myself \\
 differ& but, not, really, didnt \\
 leisure & fun, game*, relax*, family \\
 discrep & want, need, would, should \\
 sad & sad, miss, hurt*, missed \\
 risk & stop, problem*, avoid* \\
 anx & stress*, nervous, worried \\
 ingest & food*, dinner*, lunch* \\
 body & sleep, slept, stomach* \\
 insight & feel, know, think, found \\
 affiliat & we, friends, friend, love \\
 reward & good, got, get, great \\
 feel & feel, feeling, felt, hard \\
 family & family, mom, sister*, dad \\
 we & we, our, us, weve, lets\\
 \bottomrule
 \end{tabularx}
 \caption{\label{tab:liwc} The most informative LIWC features ranked by Information Gain.}
 \end{table}

We first test the separability of the positive and negative sentences with
an SVM classifier from Weka 3.8, using as baselines only unigrams and
LIWC \cite{PennebakerFrancis01} as features.  Results for these
baseline classifiers are in
Table~\ref{tab:baseline-SVM-results}, illustrating that the positive
and negative classes can be separated with F1 above .70, and that
both unigrams and LIWC perform worse on the negative
class. 

\begin{table*}[ht!]
%\begin{scriptsize}
\begin{small}
\centering
\begin{tabular}{p{.65in}|p{2in}|p{3.15in}}
\toprule
\bf Well-Being & \bf Frames & \bf Example Lexical Units \\
\midrule Goal & Desiring, Intentionally\_Act, Purpose & want, feel like, hope,
wanted, wish, do, did, done, doing, does, plan, purpose, in order, intention, goals \\ 
\midrule 
Autonomy \& Obligation & Being\_obligated, Required\_event, Avoiding, Inhibit\_movement, 
Have\_as\_requirement, Complaining & complain, grumble, complaints, have to, had to,
should, having to, need, get to, had to, have to, got to, should, avoid , ducking, take, need, needed, requires\\ 
\midrule 
Competence & Activity\_done\_state, Attempt, Capability, Bungling, Difficulty, Practice, Activity\_finish, Accomplishment & finished, trying, try, tried, effort, attempt, efforts, can, could, exercise, practice, rehearsal, exercising,
able, ability, unable, messed up; ruined; screwed up
, ruin, hard, difficult, easy, tough, easier challenging, impossible, a breeze, hardest, finish, finishing, completed, accomplished, achieve \\
\midrule
Connection \& Lack-of Connection &
Death, Forming\_relationships, Social\_event, Kinship, People, People\_by\_residence, Telling, Communication\_response & 
birthday, married, divorce, befriend, dinner,
social, party, picnic, mom, family, parents, sister, cousin, told,
tell, informed, people, girl, man, roommate, reply, answers, answer, reacted \\ \midrule

Savouring & Emotions\_of\_mental\_activity, Feeling, Annoyance, Desirability, Food, Chemical-sense
description, Ambient\_temperature, Emotions-by-stimulus, Stimulus\_focus, Intoxicants, Communication\_noise, 
Experiencer\_Focus, Perception\_experience, Biological\_urge, Death & enjoyed, like, hate, glad, annoyed, cry, yelled, whooped, honked, irritated, feel, feeling, yummy, alcohol, weed, drugs, dope, see, felt, seeing, hear, experience, 
senses, experiences, taste,  feel, 
delicious, tasty, sweet, food, coffee, bread, cheese,
good, bad, great,  better, best, horrible, worst 
wonderful, weird, nice,  relaxing, annoying, interesting,  sad, weird 
enjoyable, comforting, entertaining, unpleasant, hilarious, rest, relaxation, exhilarating,  tiring, nicer, disturbing, disappointing, embarrassing, irritating, upsetting, heartbreaking, consoling, tedious,  traumatic, chilling, calming, frightening
touching, pleasure, satisfying, fascinating, tired, exhausted, sleepy, hungry, nauseated, horny 
\\ 
\bottomrule
\end{tabular}
%\end{scriptsize}
\end{small}
\caption{\label{tab:framecat} Frame Categories and Associated Well-Being Classes. }
\end{table*}

However, as discussed above, the word level representations of the
features in the baselines do not help us with our goal to understand
how linguistic descriptions of events that affect well-being map onto
theoretical constructs.  Table~\ref{tab:unigram} and
Table~\ref{tab:liwc} provide the most informative UniGrams and LIWC
categories.  We cannot recommend to an {\sc echo} user that they should
for example, try to use the word {\it why} less (Row 7) because
it is correlated with negative feelings, or try to use less negation (Rows 9 and 10).
 It is difficult to associate these features with
well-being classes. Even in cases where the words seem to be strongly
related to a well-being category, a single word typically fails to provide enough
information, e.g., ``it was \textbf{fun}
talking to him" and ``worked on a \textbf{fun} project"  belong to
different well-being classes.  Moreover, the mapping of LIWC
categories to words are many-to-many, e.g.  the ``discrep" category
contains words related to both Goals and Autonomy. We posit that we
need compositional semantic features to ground our a Well-Being
classification of events.

We thus explore two
different methods for mapping these well-being event categories into
lexical descriptions, one of which is top-down and the other which is
bottom-up.  Our top-down method is based on mapping general event
types from FrameNet to the theoretical categories enumerated in
Table~\ref{theory-examp}. We take frame specific features for each
theoretical category from the lexical units for each frame. For
example, {\sc goals} are often discussed in terms of specific frames
from the  Desiring and the Intentionally\_act
classes, as shown in the first two rows of
Table~\ref{tab:framecat}.

We show that FrameNet features do provide an interesting level of
generalization but much of the compositional semantics of events is
still missing from this characterization (Section~\ref{frame-sec}).  Thus, our bottom-up method
applies the AutoSlog linguistic-pattern learner to induce
lexically-grounded predicate patterns from the {\sc echo} data (Section~\ref{ling-patt-sec}).
We show how many light verbs acquire a specific semantics with their
arguments, and how common events like ``Talking'' are separated
into positive and negative events depending on whether they are
``Talking about'' or ``Talking with''.

\section{Frames and Well-Being}
\label{frame-sec}

Table~\ref{tab:framecat} provides our posited mapping from frame
categories to the appraisal category of {\sc goals}
as well as to the eudaimonic categories of {\sc autonomy}, {\sc
  competence} and {\sc connection}, and to the hedonic category of
{\sc savouring}.  To develop features related to these frame
categories, we apply SEMAFOR \cite{das2013:semafor} to label the {\sc
  echo} posts with their corresponding frames using FrameNet 1.5
\cite{Bakeretal15,Bakeretal14}. We partition frame features into
subsets corresponding to the different theoretical constructs
as defined in Table~\ref{tab:framecat}. We acknowledge that our
mapping may not be perfect, and that some frames could conceivably
be categorized as both goal related and eudaimonic.

\begin{table}[h!t]
\footnotesize
\centering
\begin{tabularx}{2.8in}{p{0.85in}| c| c c  c}
\toprule
\bf Features & \bf Metric & \bf Pos & \bf Neg & \bf All  \\
\midrule 
GOALS &  Prec &  0.62  & 0.49 &  0.57  \\
        &  Rec & 0.94 & 0.09  & 0.61 \\ 
        &  F1 &  0.75 & 0.15 & 0.51 \\
\midrule 
EUDAIMONIC &  Prec & 0.63 & 0.58 & 0.61  \\
     &  Rec & 0.93 & 0.16 & 0.63  \\
     &  F1 & 0.75 & 0.25 &  0.56 \\
\midrule 
SAVOURING &  Prec & 0.61 & 0.44 & 0.55  \\
     &  Rec & 0.97 & 0.04 & 0.61  \\
     &  F1 & 0.75 & 0.08 &  0.49 \\
\midrule
ALL FRAMES &  Prec &  0.69  & 0.74 &  0.71  \\
        &  Rec & 0.91 & 0.38  & 0.70 \\ 
        &  F1 &  0.78 & 0.50 & 0.67 \\
\bottomrule
\end{tabularx}
\caption{\label{tab:SVM-results} Coverage of Different Theoretical Categories. }
\end{table}  

We train an SVM with each feature subset, and evaluate the models on
our test set, with results in Table~\ref{tab:SVM-results}. The general
{\sc all frame} feature is also listed for comparison. The .67 F1 of
{\sc frame} is slightly lower than {\sc liwc} in
Table~\ref{tab:baseline-SVM-results}, but in our view, more
interpretable.  In addition, the average count of {\sc frame} features
per sentence is an order of magnitude less than {\sc liwc} features
(hence, much less than unigram features), suggesting the targeted
power of these features.  See Table~\ref{tab:feature-counts}. We posit
that {\sc frames} are thus more discriminative than {\sc liwc} for
well-being classes, and that {\sc frame} features are
more naturally categorized into  well-being categories at a semantic
level.

\begin{table}[h!t]
\footnotesize
\centering
\begin{tabularx}{2.8in}{p{0.8in}| c| c c |  c}
\toprule
\bf Features & \bf Dataset & \bf Pos & \bf Neg & \bf Total \\
\midrule
 UniGram &  Train &  8.5  & 9.9 &  9.1  \\
         &  Test & 8.1 & 9.8  & 8.7 \\
 \midrule
 LIWC &  Train & 25.4 & 31.4 &  27.8  \\
      &  Test & 23.8 & 30.6 & 26.4  \\
 \midrule
 ALL FRAMES &  Train & 2.7 & 5.2 & 3.7  \\
      &  Test & 3.3 & 4.0 & 3.6  \\
 \bottomrule
 \end{tabularx}
 \caption{\label{tab:feature-counts} Average Feature Counts for Sentence}
 \end{table}

\begin{table}[h!t]
\footnotesize
\centering
\begin{tabularx}{2.8in}{p{0.85in}| c| c c  c}
\toprule
\bf Features & \bf Metric & \bf Pos & \bf Neg & \bf All  \\
\midrule 
AUTONOMY &  Prec &  0.0  & 0.39 &  0.15  \\
        &  Rec & 0.0 & 1.0  & 0..39\\ 
        &  F1 &  0.0 & .56 & 0.22 \\
\midrule 
COMPETENCE &  Prec & 0.56 & 0.58 & 0.60  \\
     &  Rec & 0.98 & 0.04 & 0.61  \\
     &  F1 & 0.76 & 0.07 &  0.49 \\
\midrule 
CONNECTION&  Prec & 0.62 & 0.58 & 0.60  \\
     &  Rec & 0.97 & 0.06 & 0.62  \\
     &  F1 & 0.76 & 0.11 &  0.49 \\
\bottomrule
\end{tabularx}
\caption{\label{tab:EUD-SVM-results} Results for Individual Eudaimonic Categories. }
\end{table}  

The Goals section of Table~\ref{tab:SVM-results} shows that
Appraisal theory does well at predicting positive events, but
performs poorly for negative events, primarily due to low recall. All features
achieve good F1 for the
positive class, but not the negative class. This is
consistent with the results in
Table~\ref{tab:baseline-SVM-results}.

\begin{table*}[t]
%\begin{scriptsize}
\begin{small}
\centering
\begin{tabularx}{6.3in}{p{0.85in}|p{1.4in}|p{0.35in} |c}
\toprule
\bf Well-Being &\bf  Frame & \bf Affect & \bf Example \\
\midrule
GOALS & Desiring &  POS & I think it went well and I {\bf hope} I did a good job. \\
 & Intentionally\_act & NEG & My midterm was really long and I  {\bf didn't} finish.\\
\midrule
AUTONOMY & Being\_obligated & NEG & I'm mad that I {\bf had to} drive all the way to Fresno.\\
& Required\_event & NEG & I {\bf need to} stay awake and listen, but it 's hard.\\
\midrule
COMPETENCE & Capability & POS &  I feel so empowering whenever I'm {\bf able to} help others. \\
 & Attempt &  NEG & {\bf Tried to} chat with some people online, did n't work out. \\
\midrule
CONNECTION&  Kinship & POS & My  {\bf mom} and I hung out and walked around for 6 hours .\\
  & Telling & NEG & I wonder how much they will {\bf tell} me my teeth are bad today.\\
\midrule
SAVOURING & Chemical-sense\_description& POS & {\bf Yummy} burgers and sides.\\
 & Food & POS & Made homemade ice cream with my husband:...{\bf cookie dough}\\
\bottomrule
\end{tabularx}
%\end{scriptsize}
\end{small}
\caption{\label{tab:frame-examples} Top Frame Categories and Associated Well-Being Classes. }
\end{table*}

The {\sc eudaimonic} features include Autonomy \& Obligation,
Competence and Connection. The SVM trained with just eudaimonic
features produces the highest F1 score for the negative class,
highlighting the role of eudaimonic related events in negative
well-being. See Table~\ref{tab:SVM-results}. The results for an
breaking eudaimonic into its constituent
categories is in Table~\ref{tab:EUD-SVM-results}. The results
show that most of our autonomy categories are related to negative
autonomy, to obligations that cause feelings of negative
well-being. On the other hand, the results indicate that competence
and connection play a large role in positive well-being.

The top 25 most informative frame
features are illustrated in Table~\ref{tab:frame-examples} (out of 639 instantiated
in {\sc echo}). These illustrate general events for well-being, but
compositional differences, such as
``spending my nights by the side of my textbook'' and ``spending my
nights with friends'' are not captured.  The first ``spend (time)'' evokes
the theoretical construct of  obligation, while ``spend (time with)'' is related to
connection.

\section{Linguistic Pattern Learning}
\label{ling-patt-sec}

We also apply Autoslog-TS, a weakly supervised linguistic-pattern
learner as a way of learning some compositional patterns. Autoslog
only requires training documents labeled broadly into our two classes
of {\sc positive} or {\sc negative}. The learner uses a set of
syntactic templates to define different types of linguistic
expressions.  In general, this method tends to produce high precision
(and potentially low recall) markers of the particular classes that
can seed further hypothesizing.

The left-hand side of Table~\ref{pattern-types} lists example pattern
template and the right-hand side illustrates a specific
lexico-syntactic pattern ({\bf in bold}) that represents an
instantiation of each general pattern template for learning well-being
patterns in our data.\footnote{The examples are shown as general
  expressions for readability, but the actual patterns must match the
  syntactic constraints associated with the pattern template.} 

\begin{table}[t!bh]
%\begin{scriptsize}
\begin{small}
\centering
\ra{1.3}
\begin{tabular}{@{}p{0.1in}|p{1.1in}|p{1.4in}@{}}\toprule
\bf \cellcolor[gray]{0.9} & \bf \cellcolor[gray]{0.9} Pattern Template & \bf \cellcolor[gray]{0.9} Example Instantiations \\ \midrule
1 & $<$subj$>$ PassVP & {\bf $<$I$>$ am so relaxed} after getting to sleep in and rest all morning. \\ \midrule
2 &   $<$subj$>$ ActVP        & When it does happen, {\bf $<$I$>$ feel energized} because IT IS a special experience to me.\\ \midrule
3&    $<$subj$>$ ActVP Dobj   & {\bf $<$I$>$ enjoy his efforts} lately to make me happier.\\ \midrule
4&    $<$subj$>$ ActInfVP & Found some some stuff but I AM not sure if {\bf $<$I$>$ want to keep} them.  \\ \midrule
5&    $<$subj$>$ PassInfVP & 2 of {\bf $<$my housemates$>$ were supposed to clean} on Tuesday and they still haven't. \\ \midrule 
6&    $<$subj$>$ AuxVP Dobj & We ate and {\bf $<$We$>$ had a glass of my favorite wine}.
\\ \midrule
7&    $<$subj$>$ AuxVP Adj & {\bf $<$All of the colors$>$ are so much more vibrant}. \\ \midrule \midrule
8&    ActVP $<$dobj$>$ & Cannot wait to study while {\bf eating $<$this$>$}. \\ \midrule
9&    InfVP $<$dobj$>$ & Just realized I {\bf forgot to turn in $<$my homework$>$}.\\ \midrule
10&    ActInfVP $<$dobj$>$ & I really {\bf need to start $<$my hw$>$} sooner...\\ \midrule
% 11 &   PassInfVP $<$dobj$>$ & i love it when people do this. 'you have to prove everything you say, but i {\bf am allowed to }simply {\bf make $<$assertions$>$} and it's your job to show i'm wrong.'\\ \midrule
 11&       Subj AuxVP $<$dobj$>$   & IT IS the Super Bowl today and {\bf THERE IS $<$a party$>$} at my house. \\ 
    \midrule \midrule
12&    NP Prep $<$np$>$  &  {\bf Driving in $<$the rain$>$} is scary.\\ \midrule
13&    ActVP Prep $<$np$>$ & Almost as if I forgot something terribly important or I {\bf messed up $<$something$>$} important in my life. \\ \midrule
14&    PassVP Prep $<$np$>$ & And I feel like I did but just this once I messed up and I might {\bf be punished for $<$it$>$}. \\  \midrule
15&    InfVP Prep $<$np$>$ & Felt amazing to be {\bf done with $<$finals$>$}!\\ \midrule
16&    $<$possessive$>$ NP & {\bf $<$Her$>$ attitude} is not working anymore. \\ 
 \bottomrule
 \end{tabular}
%\end{scriptsize}
\end{small}
\vspace{-.1in}
  \caption{\label{pattern-types} AutoSlog-TS Templates and Example Instantiations}
\vspace{-.1in}
\end{table}

\begin{table*}[t!h]
%\begin{scriptsize}
\begin{small}
\ra{1.3}
\begin{tabular}{@{}P{0.55cm}|P{0.52cm}|P{5.9cm}|P{7.7cm}@{}}
\toprule
\bf Prob. & \bf Freq. & {\bf Pattern and Text Match} & \textbf{Sample Post}\\ \hline
\multicolumn{4}{c}{\bf \cellcolor[gray]{0.9}  Positive Example Patterns}  \\\hline
\bf 1.00 & 11 & ActVp Prep $<$NP$>$ (\texttt{WENT ON}) &  I just  \textbf{went on} a hike this is the best thing ever. \\ 
\bf 1.00 & 7 & $<$subj$>$ ActVP Dobj (\texttt{MADE FOOD}) &  \textbf{Made a German pancake} for breakfast. \\ 
\bf 1.00 & 7 & NP Prep $<$np$>$  (\texttt{CATCHING WITH}) &  \textbf{Catching up with} old friends! \\ 
\bf 1.00 & 7 & ActVP $<$dobj$>$  (\texttt{USED}) &  \textbf{Used} the Laurel's Kitchen Bread Book recipe. \\ 
\bf 1.00 & 6 & ActVP Prep $<$np$>$  (\texttt{GOT OFF}) &  \textbf{Got off} work.\\ 
\bf 1.00 & 4 & NP Prep $<$np$>$  (\texttt{TALK WITH}) &  Having a really nice \textbf{talk with} my aunt.\\ 
\bf 0.95 & 18 & ActVP $<$dobj$>$  (\texttt{FINISHED}) & \textbf{Finished} my paper.\\ 
\bf 0.78 & 39 &ActVP $<$dobj$>$  (\texttt{TOOK}) & \textbf{Took} a walk after class and truly enjoyed the outdoors! \\ 
\bf 0.78 & 25 & $<$subj$>$ ActVP (\texttt{ATE}) & We  \textbf{ate} and had a glass of my favorite wine. \\ 
\bf 0.73 & 11 & InfVP Prep $<$np$>$  (\texttt{SPEND WITH}) &  Happy to simply \textbf{spend time with} friends.\\ 
\hline
\multicolumn{4}{c}{\bf \cellcolor[gray]{0.9}  Negative  Example Patterns}  \\ \hline

\bf 1.00 & 9 & InfVP $<$dobj$>$ (\texttt{AVOID}) & Better buy ... in smaller packaging to \textbf{avoid} wasting again.\\ 
\bf 1.00 & 8 & ActVP $<$dobj$>$ (\texttt{USE}) & All she did was \textbf{use} water and wipe a few corners. \\ 
\bf 1.00 & 7& InfVP $<$dobj$>$  (\texttt{STOP}) & I need to \textbf{stop} smoking. \\ 
\bf 1.00 & 6 & $<$subj$>$ ActVP Prep $<$np$>$ (\texttt{NOT TALK TO}) & And now my bf is busy and \textbf{can't talk to} me. \\ 
\bf 1.00 & 5 & $<$subj$>$ ActVP Dobj  (\texttt{TEXTED ME}) & He \textbf{texted me} finally but then he randomly stopped. \\ 
\bf 1.00 & 5 & $<$subj$>$ ActVP (\texttt{NOT SLEEP}) & Have to get up early and I \textbf{can't sleep}.\\ 
\bf 1.00 & 4 & ActVP $<$dobj$>$ (\texttt{NOT FIND}) & I \textbf{did not find} the time to finish my homework. \\ 
\bf 0.82 & 14 & $<$subj$>$ ActVP (\texttt{REALIZED}) & I JUST \textbf{realized} that I have to go tomorrow. \\ 
\bf 0.81 & 13 & $<$subj$>$ ActVP (\texttt{TAKE}) & Since I \textbf{take} around 35 minutes to get ready, I missed ... \\ 
\bf 0.80 & 20 & ActVP $<$dobj$>$ (\texttt{TOLD}) & \textbf{Told} my mom about my grades. \\ 
 \bottomrule

 \end{tabular}
\end{small}
%\end{scriptsize}
\caption{\label{table:high-patterns} Examples of Characteristic ECHO Patterns using AutoSlog-TS Templates\\}
\vspace{-.1in}
\end{table*}

In order to enable selection of particular patterns,
AutoSlog-TS computes statistics on
the strength of association of each pattern with each class,
i.e. P({\sc positive} $\mid$ $p$) and P({\sc negative} $\mid$
$p$), along with the pattern's overall frequency. We define two tuning parameters for each
class: $\theta_f$, the frequency with which a pattern occurs,
$\theta_p$, the probability with which a pattern is associated with
the given class.

AutoSlog lets us systematically explore tradeoffs with precision and recall.
Here we select $\theta_f$ and $\theta_p$ to optimize F1 on our test set.
For more detail, see \cite{Riloff96,Orabyetal15}. 

Our primary interest here is Autoslog's ability to learn compositional
patterns.  Autoslog can, in principle, provide three kinds of
information: i) it can provide supplement the lexical units for a
given frame; ii) it can supplement the frames in a well-being
category; and iii) it can reveal reliable markers of mood that
well-being categories do not capture. Because our interest in frames
is ultimately as a way of relating well-being categories with
linguistic signals, we will not distinguish (i) and (ii) here.

Here we discuss all patterns with a $\theta_p>.7$ Several
lexicosyntactic patterns fit within our well-being categories but are
not captured by frames, while as expected there are overlaps between
FrameNet and Autoslog as well.  Examples are listed in
Table~\ref{table:high-patterns}.  One large class includes
straightforward lexical patterns: {\sc finished, finish}, and {\sc finally}
which we associate with feelings of comptence.
Verbal patterns with {\sc eat} and {\sc ate} indicate savouring, with
{\sc not\_eat} reliably marking negative sentences.

The frames also show many specific types of food (cake), and we use
a comprehensive list from DBpedia \cite{Lehmann2014} to collapse all these
to the general type \textsc{food}, allowing us to develop patterns such as {\sc made\_food}.

Autoslog also discovers many patterns syntactically linking content
(nouns and verbs) and function words (e.g., prepositions and light
verbs). It thus furnishes a ready source for multi-word, partially
compositional expressions of positivity or negativity. In what follows,
we provide some examples (note that in the patterns below, expressions in 
brackets are used to indicate expressions not part of the pattern that correlate
with it in the data). 

There are 262 positive patterns of the form Verb/Noun + ``with'', e.g.
{\sc talked\_with}, {\sc dinner\_with}, {\sc breakfast\_with}, {\sc
  studying\_with}, {\sc played\_with}, {\sc time\_with}, {\sc
  met\_with}, {\sc shopping\_with}, {\sc coffee\_with}, all of which
describe activities that involve connection. There are
also 100 negative patterns of this form, which are much more heterogenous,
involving both negative social experiences ({\sc argument\_with}, {\sc drama\_with},
\textsc{infuriated\_with}, \textsc{fight\_with}), but also various problematic events
(\textsc{stressed\_with}, \textsc{difficulties\_with}, \textsc{dissatisfied\_with}) and instruments
for negative events (\textsc{stop\_with}, \textsc{poisoning\_with}). Moreover, while the
positive patterns cover 523 sentences in the data, the negative
patterns cover only 133.

There are 62 patterns involving the string ``talk", 32 positive (71
items) and 30 negative (66 items).  The positive ones strongly
indicate connection (e.g., \textsc{talk\_with}, \textsc{have\_talk},
\textsc{remember\_[to]\_talk}, \textsc{got\_[to]\_talk}, \textsc{talk\_through}).
In contrast, the negative index either
the obligation to talk (e.g., \textsc{trying\_talk}, \textsc{need\_talk}, \textsc{have\_[to]\_talk})
or a failure to talk (e.g., \textsc{not\_talk\_to}, \textsc{not\_want\_talk}, \textsc{stop\_talking}).

There are 36 patterns with the string `go', 12 positive (16 items) and
24 negative (40 items).  There are 34 patterns involving the past
tense form ``went'', which reverses the polarity to 25 positive
patterns (273 items) and 9 negative (9 items).  Across the two
versions of the lemma, the positive patterns provide several
expressions for savouring (\textsc{went/go\_on/for} [a walk, a hike, a ride],
\textsc{went/go\_shopping/swimming}, \textsc{went/go\_to} [the mall, a movie]).  For the
negative, the predominance of `go' comes from the fact that they are
largely negated (\textsc{not\_go\_to} [the movies]) or in infinitive contexts
that suggest obligation (\textsc{[have to]\_go\_to} [class], \textsc{[have to]\_go\_work}).
Similarly, the positive class contains 9 patterns with
`bought' and 1 with `buy' (\textsc{enticed\_[to]\_buy}) and the negative
class has 6 patterns with `bought' and 16 with `buy', all emphasizing
buying necessities (\textsc{buy\_groceries/ticket}, \textsc{need/want\_buy}, \textsc{not\_buy})
Thus, even though these expressions all involve the same verbs and
prepositions, the surrounding environments, as reflected in the form
of the verb, split between positive and negative sentence classes.

There are 73 bigram patterns of the form \textsc{new\_x}, 56 positive (83 items) and 17
negative (21 items).  In general, the positive ones describe new
objects -- \textsc{shirt}, \textsc{sheets}, \textsc{computer}, \textsc{clothes}, \textsc{tea} -- and
acquaintances (\textsc{new\_friend}), thus encompassing both Connection and
possibly Savouring.  In contrast, the negative patterns describe
changes to routines -- \textsc{habits}, school \textsc{quarter}, \textsc{professor},
\textsc{living} [conditions], or \textsc{schedule} -- which are likely to engender a
sense of instability, and hence be Eudaimonically negative.

Thus, these patterns illustrate that Autoslog can serve as a high-precision
method of building additional patterns -- especially compositional ones -- 
for a given well-being category.

\section{Conclusions and Future Work}
In this paper, we have advanced a synthetic categorization of the
sources for well-being and happiness. We have used a corpus of private
micro-blogs from the {\sc echo} application to explore how well we can
map linguistic expressions of well-being to this classification. We
have shown that FrameNet provides useful generalizations, while the
linguistic pattern learner AutoSlog illustrates the details and
challenges of the compositional nature of user's descriptions of their
daily experiences. Moreover, we have demonstrated that, independently,
each of these methods can produce performance similar to that of
conventional lexical methods with a feature space that is smaller,
and, in the case of FrameNet features, psychologically grounded. Our
Autoslog exploration moreover reveals a way of exploring the space of
patterns that our FrameNet mapping has missed. In future work, we aim
to automatically combine these two methods and bring the Autoslog
patterns under the well-being categorization we have advocated here.
We also plan to investigate new models with the untouched 6224 Echo
posts, as well as larger public corpus like LiveJournal.

In addition, we plan to explore the source of the fact that there are
more positive patterns (both as types and the tokens they capture)
than the negative ones, which directly relates to the lower Neg recall
for all classifiers we tested.  While we could not find any clear
reason in our examination of the data, this asymmetry may indicate
that markers of negativity are more syntactically distributed than our
current list of patterns looks for, or perhaps less linguistically
reliable.

\section*{Acknowledgments}

This research was partially supported by NSF Robust Intelligence \#IIS-1302668-002 and 
NSF HCC \#IIS-1321102.

\bibliographystyle{emnlp_natbib} 

\bibliography{nl}
%\bibliography{../../nl}

%\bibliographystyle{plain}

\end{document}